\documentclass[10pt, a4paper]{article}

\usepackage{lrec-coling2024} 
\usepackage{multirow}

\usepackage{amsmath}

\usepackage{authblk}

\title{Generating Clarification Questions for Disambiguating Contracts}

\name{Anmol Singhal, Chirag Jain, Preethu Rose Anish \\ 
{\bf \large Arkajyoti Chakraborty* \thanks{* Intern at TCS Research (Feb-July 2023)}, Smita Ghaisas}}

\address{TCS Research, India \\
         \{anmol.singhal, chirag.rjain3, preethu.rose, smita.ghaisas\}@tcs.com\\
         arkajyoti09062001@gmail.com \\
         }

\abstract{
Enterprises frequently enter into commercial contracts that can serve as vital sources of project-specific requirements. Contractual clauses are obligatory, and the requirements derived from contracts can detail the downstream implementation activities that non-legal stakeholders, including requirement analysts, engineers, and delivery personnel, need to conduct. However, comprehending contracts is cognitively demanding and error-prone for such stakeholders due to the extensive use of Legalese and the inherent complexity of contract language. Furthermore, contracts often contain ambiguously worded clauses to ensure comprehensive coverage. In contrast, non-legal stakeholders require a detailed and unambiguous comprehension of contractual clauses to craft actionable requirements. In this work, we introduce a novel legal NLP task that involves generating clarification questions for contracts. These questions aim to identify contract ambiguities on a document level, thereby assisting non-legal stakeholders in obtaining the necessary details for eliciting requirements. This task is challenged by three core issues: (1) data availability, (2) the length and unstructured nature of contracts, and (3) the complexity of legal text. To address these issues, we propose ConRAP, a retrieval-augmented prompting framework for generating clarification questions to disambiguate contractual text. Experiments conducted on contracts sourced from the publicly available CUAD dataset show that ConRAP with ChatGPT can detect ambiguities with an F2 score of 0.87. 70\% of the generated clarification questions are deemed useful by human evaluators. 
 \\ \newline  }

\begin{document}

\maketitleabstract

\section{Introduction}

Contracts are legally binding agreements among parties involved, with non-compliance often resulting in severe penalties. These documents serve as indispensable sources for enterprises to gather project-specific requirements for meeting customer demands \cite{Soavi2022-tc, smita-1}. The requirements derived from contracts are necessary to outline the scope of downstream implementation activities, which include designing the solution, engineering components of the solution, maintaining security and privacy barriers, and project management tasks related to delivery and intellectual property. However, contracts are confidential documents and are often only accessible to the top management and compliance teams of enterprises \cite{smita-2}. Directors and project managers typically assign contractual clauses to specific teams of non-legal stakeholders for requirements elicitation and subsequent implementation \cite{singhal-etal-2023-towards}. These teams encompass requirement analysts, engineers, designers, and delivery personnel, who are vital to translating contractual obligations into tangible actions. However, deciphering contractual clauses to extract relevant requirements is error-prone and imposes a significant cognitive burden. The task is challenging because non-legal stakeholders do not have a fluent understanding of the Legalese used in contract drafting. Moreover, many clauses in standard-form contracts are drafted with ambiguity to allow for flexibility and adaptability in interpreting their terms \cite{LI201798}. In contrast, non-legal stakeholders require a detailed understanding of the clauses to formulate actionable requirements. Ambiguous clauses can negatively impact downstream implementation activities and instigate disputes among the involved parties, potentially leading to substantial financial and legal ramifications.

The introduction of an AI-enabled assistant aimed at contract clause disambiguation can be an effective solution. The assistant can automatically recommend Clarification Questions (CQs) that non-legal stakeholders should ask customers or their representatives to resolve ambiguities. The traditional question-generation task in NLP relies on the existence of answers in the given context to formulate questions. However, CQs are designed to pinpoint missing or ambiguous information within the context. Providing AI-generated CQs can serve as a check for non-legal stakeholders to ensure that contractual clauses are precise and clear for implementation. Therefore, CQs can significantly decrease the human effort associated with requirements elicitation.

While previous research has mainly focused on Clarification Question Generation (CQGen) for disambiguating conversational search queries and conversational Question Answering (QA) on e-commerce portals \cite{rahmani2023survey}, to the best of our knowledge, there has been no exploration of CQGen application in contracts and legal text. Furthermore, there is no prior work on detecting ambiguities in legal text on a document level. Existing techniques assume the presence of labeled data and focus on domains that require limited context to determine ambiguities. Contractual application of CQGen presents unique challenges: (1) the unavailability of labeled datasets, (2) the length and unstructured nature of contract documents, and (3) the complex language commonly used in contract drafting \cite{xu2022conreader}. 

A promising direction to address these challenges is to utilize the rapidly advancing capabilities of Large Language Models (LLMs) and directly prompt them to generate CQs. Chain-of-Thought (CoT) prompting \cite{wei2023chainofthought}, which encourages the model to reason step-by-step before producing the answer, has shown potential on several reasoning tasks. However, it is not well-suited for reasoning over lengthy documents such as contracts because obtaining the output of intermediate steps is non-trivial \cite{sun2023pearl}. 

We propose \textbf{ConRAP}, a \textbf{Con}tract-specific \textbf{R}etrieval \textbf{A}ugmented \textbf{P}rompting framework for generating CQs to address these core challenges. ConRAP takes as input one contractual sentence at a time and uses a novel prompting technique, termed ‘Attribute Prompting', to detect (1) if the given sentence is ambiguous and (2) generate a set of candidate CQs highlighting the ambiguities within the sentence. Subsequently, ConRAP searches for answers to each generated CQ within the overall contract document using a retrieval-augmented QA approach. If the contract already addresses the CQ, it is filtered out from the candidate set, and the QA pair is returned to the user for reference. The remaining CQs in the candidate set are separately returned to the user as unanswered questions, which pinpoint ambiguities in the contract and can be raised by non-legal stakeholders to seek clarification from customer representatives. 

We evaluate ConRAP utilizing a corpus of publicly available English contracts \cite{cuad} to ascertain its accuracy in detecting ambiguities. For our evaluation, we annotate this dataset with binary labels, indicating whether the sentences are ambiguous or not. We then benchmark this dataset against various retrieval-augmented, zero-shot prompting techniques. Furthermore, we assess the overall quality of the CQs generated using the ConRAP framework across different LLMs. Our findings indicate that ConRAP significantly surpasses strong baselines, establishing its superior performance in ambiguity detection and CQGen.

The contributions of our work are as follows:
\begin{itemize}
\small
\itemsep0em 
    \item We propose ConRAP, a novel framework to detect contract ambiguities on a document level and generate clarification questions. 
    \item We release a corpus of 1000 manually annotated contractual sentences for document-level ambiguity detection. 
    \item We empirically evaluate the efficacy of ConRAP using different LLMs and prompting techniques. 
    \item We manually evaluate the usefulness of generated CQs using a set of pre-defined parameters. 
\end{itemize}

The rest of the paper is organized as follows: Section \ref{s2} describes the necessary background on ambiguity detection and CQGen; Section \ref{s3} presents the ConRAP framework; Section \ref{s4} details the experimental setup; Section \ref{s5} discusses the results obtained; Section \ref{s6} details the related work; and Section \ref{s7} concludes the paper. 

\vspace{-3mm}

\section{Background}
\label{s2}
\vspace{-1mm}
\subsection{Ambiguities in Legal Text}
\label{s2.1}
Black's Law Dictionary \cite{black-dic} defines a contract provision as ambiguous if it can be reasonably interpreted or constructed in more than one way. Although some contractual clauses are intentionally drafted with ambiguity to allow for flexibility \cite{LI201798}, it is imperative from an implementation standpoint that all non-legal stakeholders share a unified understanding of ambiguous terms. For instance, software engineers tasked with implementing solutions per contractual obligations need a clear understanding of these obligations, as it can delineate the scope of implementation. Hence, clarifying ambiguous sentences with relevant stakeholders is crucial before implementation. It can allow the non-legal stakeholders to eliminate potential misunderstandings arising from ambiguities. 
\begin{figure*}[!ht]
\centering
\includegraphics[scale=0.5]{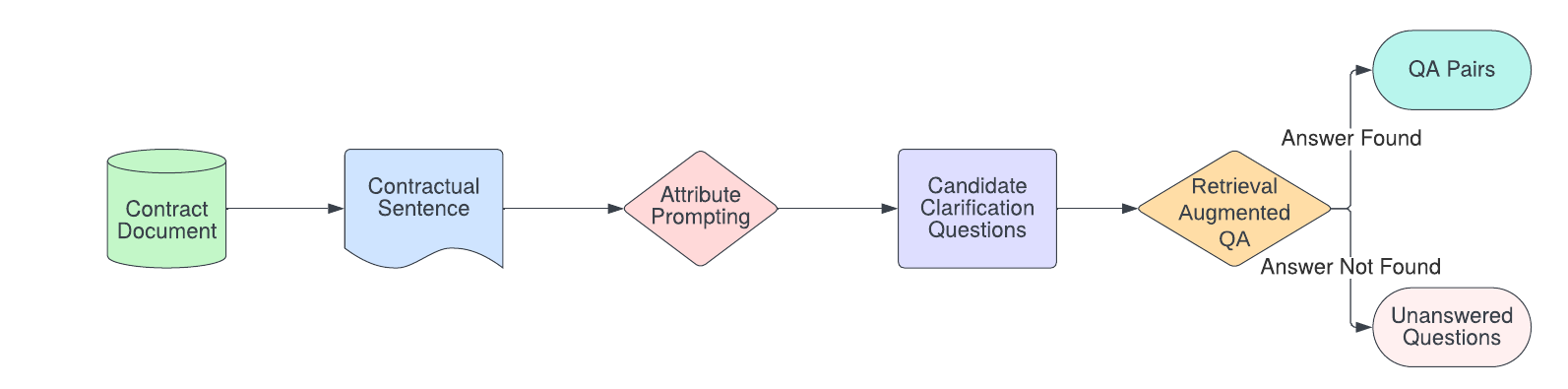}
\vspace{-3mm}
\caption{ConRAP Framework}
\label{fig1}
\vspace{-5mm}
\end{figure*}
\vspace{-1mm}

\citet{massey2014} categorized ambiguities in legal text into six types: lexical, syntactic, semantic, vagueness, incompleteness, and referential. In this research, however, our focus is confined to vagueness, incompleteness, and referential ambiguities. They can be defined as follows: 

\begin{itemize}
\vspace{-2mm}
\itemsep0em 
\small
    \item Vagueness: When a sentence contains phrases that admit borderline cases, and thus have no clear boundaries of meaning. 
    \item Incompleteness: When a sentence doesn't contain the details necessary for a precise interpretation.
    \item Referential: When a sentence refers to something that is not clearly specified.
\end{itemize}
\vspace{-1mm}
Examples of different ambiguity types in contractual sentences are provided in Appendix \ref{AppendixA}. The selection of ambiguity types is based on our initial data labeling process, during which we found no instances of lexical, syntactic, or semantic ambiguities within our dataset. The labeling process is further elaborated in Section \ref{s4.1}. 
\vspace{-3mm}

\subsection{Clarification Question Generation}
\label{s2.2}
Given a contractual sentence \begin{math}x_1\end{math} extracted from a contract \begin{math}c\end{math}, the goal of CQGen is to formulate a series of unique and semantically diverse CQs—\begin{math} <q_1,...,q_n> \end{math}—to highlight the sentence's ambiguities. To accomplish this, we adopted the formal definition of CQGen from \cite{Zhang_2021}, who presented an approach to generate a group of questions for each sentence. 

This work differentiates between the local and global contexts required for generating CQs from contracts. The local context refers to the specific sentence from which the CQ is generated. In contrast, the global context refers to the overall contract document from which the individual sentence is derived. This distinction is essential as contracts frequently involve cross-referencing, with one clause often relying on another for comprehensive interpretation. While vague phrases can be identified simply by analyzing the sentence, determining incompleteness and referential ambiguities necessitates a broader view, i.e., referencing the entire contract. Our goal is to perform ambiguity detection on a document level. Therefore, our input comprises both the sentence and the overall contract document, while the output is a group of CQs about that particular sentence.
\section{The ConRAP Framework}
\label{s3}
As discussed before, existing techniques cannot address the unique challenges involved in CQGen for contracts. ConRAP tackles each challenge as follows: (1) It incorporates a prompting-based approach to eliminate the need for a large labeled dataset for fine-tuning LLMs. (2) It uses an LLM-driven pipeline and includes attribute prompting (Section \ref{s3.1}) to mitigate the challenge posed by the complexity of legal text. (3) It deals with the length of contract documents by combining question generation and answering to optimize ambiguity detection on a document level (Section \ref{s3.2}). In this section, we elaborate on the specifics of the ConRAP framework, which incorporates two modules that separately account for local and global contextual information. The interplay of these two modules and the workflow of the ConRAP framework is depicted in Figure \ref{fig1}.
\vspace{-3mm}
\subsection{Attribute Prompting}
\label{s3.1}
\begin{figure*}[!ht]
\centering
\includegraphics[scale=0.5]{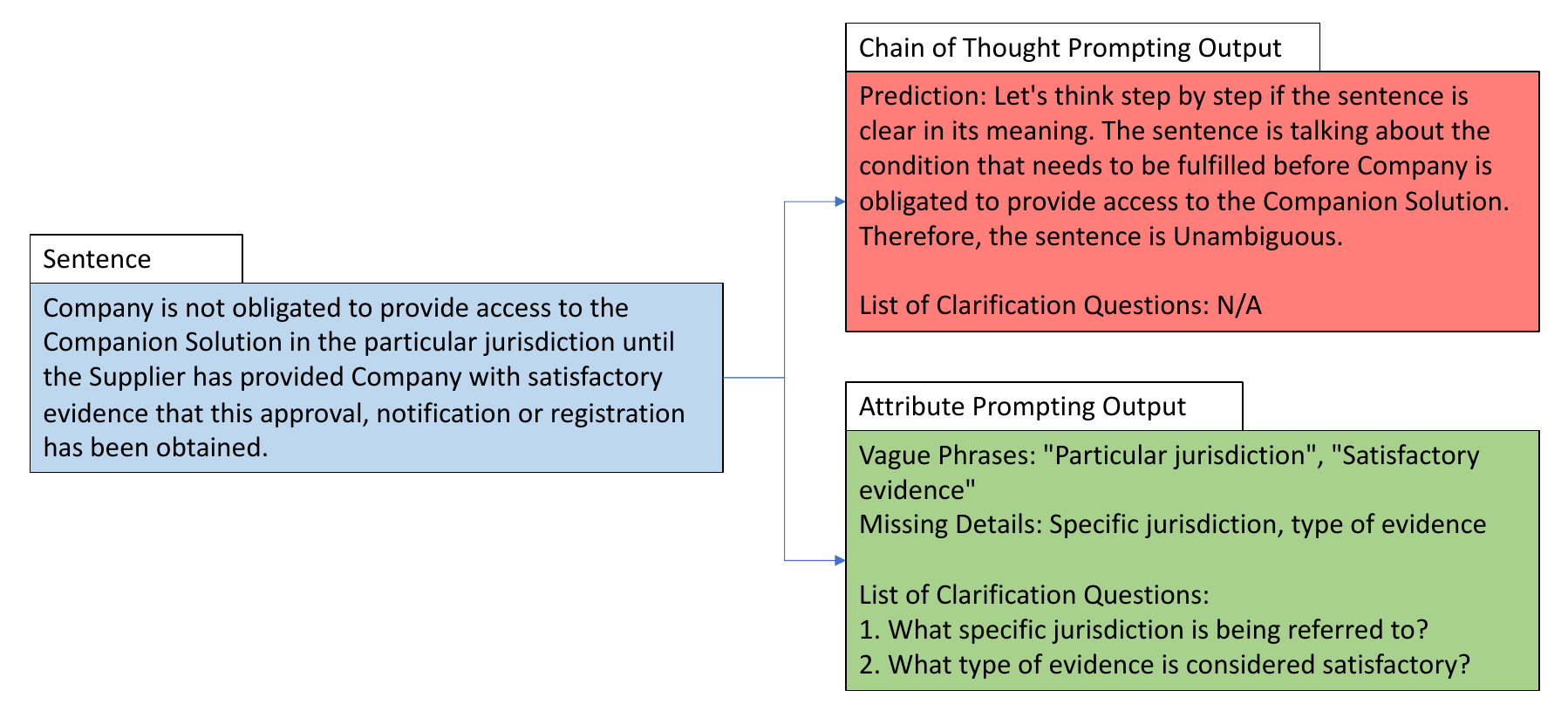}
\vspace{-3mm}
\caption{Attribute Prompting enables LLMs to detect ambiguities more comprehensively than CoT}
\label{fig2}
\vspace{-4mm}
\end{figure*}

Existing prompting techniques such as CoT are fundamentally anchored on the principle of giving the model `time to think'. This principle involves breaking down the overall task into a sequence of intermediate steps, subsequently enhancing the reasoning capabilities of LLMs beyond those offered by direct prompting. However, these techniques often fail to comprehensively detect ambiguities, an aspect we further elucidate in Section \ref{s5.1}.

Our objective with CQGen is to comprehensively identify all existing ambiguities in a contractual sentence and generate CQs based on these ambiguous terms. To address this, we propose `Attribute Prompting'. Attributes can be defined as semantic textual elements that typically occur within sentences and are vital for the precise interpretation of sentences. In our case, the sentences are from clauses present in contracts. The list of different types of attributes present in our dataset is provided in Appendix \ref{AppendixB}.

Attribute prompting generates CQs based on attributes, which, when vague, incomplete, or absent, can be a source of ambiguity within the contractual sentence. We proceed under the assumption that each attribute is independent, implying that any vague, incomplete, or missing attribute contributes uniformly to the overall ambiguity of a sentence. Therefore, we do not consider one attribute more important than another for determining ambiguities. The study of how the confluence of various attributes influences the degree of ambiguity is a subject for future work.

With this assumption in mind, we now formally describe attribute prompting. Given a sentence \begin{math}
    x
\end{math}, the model first identifies the set $A$ consisting of $m$ attributes \begin{math}
    {a_1, a_2, \text{..} ,a_m}
\end{math} within the sentence that are vague, incomplete, or absent. The LLM then generates a candidate CQ set \begin{math}
    C= \{cq_1, \text{..}, cq_m\}
\end{math} to clarify each identified attribute. 

If the LLM cannot identify any attributes that may be vague, incomplete, or absent, the corresponding sentence is marked as `unambiguous', and no CQs are generated. Conversely, if ambiguities are identified, the sentence is marked as `ambiguous', this classification being reliant on the local context. 

The prompt design used in this work consisted of four components: (1) system message that sets the context in which the LLM should operate, (2) task description, (3) output format, and (4) input sentence. We incorporated these components after referring to the various prompt engineering practices recommended in the Langchain documentation. However, we also experimented by individually adding or removing each component. The best-performing prompt included all four components. Figure  \ref{fig2} compares the output of attribute prompting with CoT. The final prompt template used for our experiments was as follows:

\noindent\fbox{%
\small
\renewcommand{\arraystretch}{1.4}
    \parbox{7.5cm}{%
\textbf{System Message}: You are a legal assistant tasked with analyzing contractual sentences for ambiguities. \\
\textbf{Task}: Given a contractual sentence, identify all vague phrases present in the sentence. Also, identify all attributes missing from the context of the sentence. \\
Generate a list of clarification questions for each vague phrase and missing detail identified. \\
\textbf{Output Format} should be as follows: \\
Vague Phrases: \textit{<Comma-seperated List>} \\
Missing Details: \textit{<Comma-seperated List>} \\
List of Clarification Questions: \textit{<Numbered List>} \\
\textbf{Given Sentence}: \textit{<input sentence>}
    }%
}

\vspace{-1mm}
\subsection{Retrieval-Augmented QA}
\label{s3.2}
Attribute prompting generates CQs purely predicated on the provided local context. However, as previously noted, contracts are intricate documents often containing cross-references. Furthermore, each provision within a contract may encompass multiple paragraphs, making a singular sentence view insufficient for accurately determining ambiguities. ConRAP accommodates this expansive global context by prompting the underlying LLM to perform retrieval-augmented QA. This module comprises two primary stages:

\textit{Contract Clause Retriever}: The goal of the retriever is to transform the raw text into a mathematical representation that can be efficiently stored and retrieved from a database. This step is challenging because most LLMs support a limited context length, and we can only pass part of the contract text as input to the LLM at a time. To overcome this challenge, we extract raw contract text from our dataset and partition it into small chunks of a pre-specified context length. During this process, we maintain an intersection window between two consecutive chunks to ascertain that no critical information is lost while splitting the text. Upon obtaining these chunks, we translate them into mathematical vectors utilizing an embedding model. These vectors are then added to a vector database that acts as the ConRAP's retriever component.

\textit{Question Answering}: In this step, ConRAP employs the vector database as an external tool for retrieving answers to the candidate CQs. Initially, it translates each CQ from the candidate set into a query vector. It then extracts the top k vectors from the index that exhibit the highest similarity to the query vector, potentially containing the answer to the query. In our study, we employed cosine similarity score as the metric for retrieval. Once the vectors most closely aligned with the CQ have been retrieved from the database, we prompt the LLM to respond to the CQ using the procured contract chunks as the relevant global context. If the LLM fails to produce an answer, we prompt it to indicate that the answer is unspecified. The prompt used for QA is given in Appendix \ref{AppendixC}. 

\begin{figure}[!t]
\centering
\includegraphics[scale=0.6]{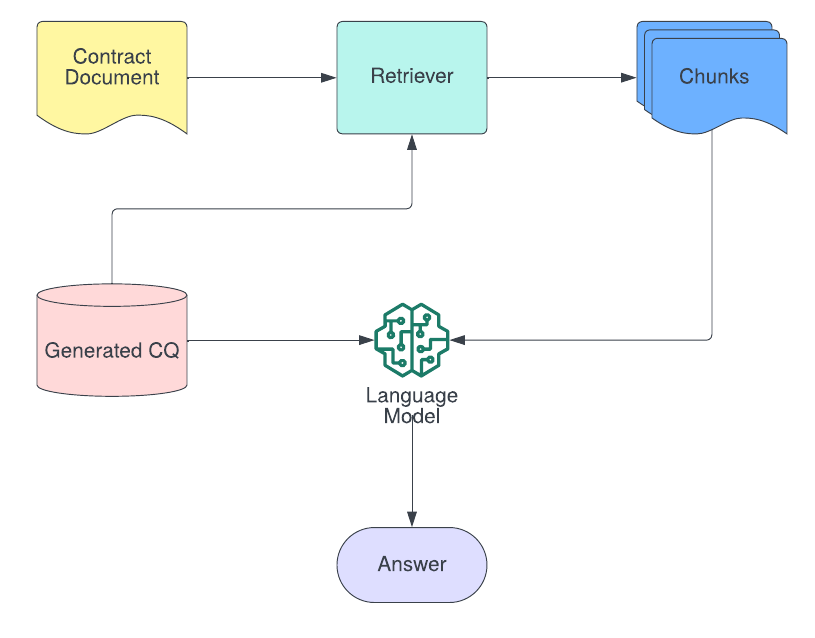}
\vspace{-3mm}
\caption{Retrieval-QA Flow in ConRAP}
\label{fig3}
\vspace{-5mm}
\end{figure}

The retrieval-QA module is illustrated in Figure \ref{fig3}. Thus, we can pinpoint CQs for which answers already exist within the contract by utilizing the retrieval-augmented QA module. These CQs are filtered from the candidate set, and the retrieved question-answer pairs are returned to the user for reference. The remaining CQs within the candidate set are designated as unanswered CQs. The sentence is deemed ambiguous if unanswered CQs remain after this stage. If all the generated CQs are filtered, the sentence is unambiguous.

\vspace{-2mm}
\section{Experimental Setup}
\label{s4}
\subsection{Dataset Creation}
\label{s4.1}
We used the Contracts Understanding Atticus Dataset (CUAD) \citelanguageresource{cuad-lr} to evaluate the ConRAP framework. It consists of 510 commercial contracts sourced from EDGAR, the Electronic Data Gathering, Analysis, and Retrieval system used by the U.S. SEC. To ensure the representativeness of our data, we used 25 different types of agreements pertaining to varied domains. We pre-processed the raw contract text by removing headings, bullets, and other special characters initially present in the text files corresponding to each contract and extracted individual sentences from contract provisions.

We delegated the labeling to four non-legal stakeholders within a large IT organization (henceforth called annotators). Each annotator had more than five years of experience in implementing various project requirements derived from contracts. Owing to the convoluted nature of legal text, the manual labeling of each sentence for all potential CQs entails a substantial human effort. Moreover, the labeling task becomes increasingly daunting when incorporating global context to discern the ambiguity of a sentence. Considering these challenges, we randomly selected 1000 sentences from the dataset and instructed the annotators to assign a binary label to each sentence- `Ambiguous' or `Unambiguous'. Since the task was to assign binary labels, each annotator assigned only one label to the sentence. We first asked the annotators to refer to the global context and analyze whether the keywords or references within a sentence were defined elsewhere in the contract from which the sentence was derived. Subsequently, we asked them to ascertain if the sentence encompassed vague, incomplete, or missing attributes. To mitigate subjectivity in labeling, we instructed the annotators to disregard the degree of ambiguity as a criterion for labeling.

Every sentence underwent independent annotation by two annotators. In total, the annotators dedicated ten working days to the annotation exercise, expending an average of four hours daily. Out of the 1000 labeled sentences, there were 120 where the annotators assigned different labels to the same sentence. The average inter-annotator agreement for assigning the labels, calculated using Cohen's kappa was 0.61. In cases of discrepancies, the sentence was forwarded to a third annotator to assign the final label.  Finally, annotators identified 524 out of 1000 sentences as ambiguous in our dataset. Table \ref{table:examples} provides examples of some contractual sentences labeled by the annotators. 

\begin{table*}
\renewcommand{\arraystretch}{1.2}
\scriptsize
\centering
\begin{tabular}{ p{2.5cm} p{4.8cm} p{2cm}  p{4.8cm}  }
\hline
\textbf{Contract} & \textbf{Sentence} & \textbf{Assigned Label} & \textbf{Explanation} \\
\hline
Embark Co-Branding Agreement & Snap will edit or write these articles as necessary to fit the overall tone of the site. & Ambiguous & The agreement does not specify the criteria that should be used by Snap to determine if editing/writing is necessary. It also does not define the site's overall tone. \\
\hline
RangeResources Transportation Agreement & Either Party may pledge this Agreement to secure any credit facility or indebtedness of such Party or its Affiliates without the consent of the other Party. & Not Ambiguous & The agreement clearly specifies the right of one party to pledge the agreement to secure a credit facility. `Affiliates' was found to be a defined term in the agreement. \\
\hline
\end{tabular}
\vspace{-2mm}
\caption{Examples of Labeled Contractual Sentences in our Dataset}
\label{table:examples}
\vspace{-5mm}
\end{table*}
\vspace{-2mm}
\subsection{Research Questions and Baselines}
\vspace{-1mm}
\label{s4.2}
We design our experiments to answer the following Research Questions (RQs):

\textbf{RQ1}: \textit{How well does ConRAP's attribute prompting detect if a sentence is ambiguous compared to other prompting techniques?}

To answer RQ1, we examine the effectiveness of attribute prompting against the following prompting techniques: (1) Direct Prompting,  where we prompt the model to directly predict if the contractual sentence is ambiguous and generate CQs, (2) CoT Prompting \cite{kojima2023large} where we prompt the model to `reason step by step' if a sentence is ambiguous and generate CQs accordingly, and (3) a modified version of CoT Prompting, where we augment the CoT prompt with additional instructions about the types of ambiguities and how it should focus on identifying minor ambiguities. The prompts corresponding to each baseline are provided in \ref{AppendixD}.

\textbf{RQ2}: \textit{Does the separation between local and global context boost ConRAP's results?}

Unlike ConRAP, existing CQGen techniques first determine the need for clarification by incorporating the relevant context (local and global) before generating CQs. Through RQ2, we examine if dealing with the local and global context separately boosts ambiguity detection results. We answer RQ2 by comparing ConRAP with the following retrieval-augmented prompting baselines:

(1) Similarity-based Retrieval: We first retrieve other clauses from the contract similar to the given sentence. The attribute prompt is then augmented with the retrieved top-k similar clauses, and the LLM is prompted to generate CQs considering both the sentence and the additional context,

(2) NER-based Retrieval: We first extract Named Entity Recognition (NER) tags from each sentence and retrieve the definition/description for each tag as mentioned in the contract document. The attribute prompt is then augmented with the extracted definitions, and the LLM is prompted to generate CQs by considering both the sentence and the relevant definitions.

\textbf{RQ3}: \textit{How accurate is ConRAP’s retrieval-augmented QA module?}

We answer this question by validating the responses obtained against the best-performing techniques based on the results of RQ1 and RQ2. The metrics for validation are specified in Section \ref{s4.3}

\textbf{RQ4}: \textit{How does the generation quality of CQs differ when using ConRAP with different LLMs?}

We answer RQ4 by comparing ConRAP’s generation quality across the following instruction-tuned LLMs: (1) ChatGPT, (2) Alpaca-LORA (7B) \cite{alpaca, hu2021lora} (3) Vicuna (13B) \cite{vicuna2023}, and (4) Dolly-V2 (12B) \cite{DatabricksBlog2023DollyV2}. While Alpaca and Vicuna are open-source under a non-commercial license, Dolly-V2 is free for commercial use.
\vspace{-3mm}
\subsection{Evaluation Metrics}
\vspace{-1mm}
\label{s4.3}
We answer the RQs by evaluating the following: (1) Ambiguity detection for the dataset described in Section \ref{s4.1} against all baselines, and (2) Quality of the generated CQs. 

\textit{Ambiguity Detection Metrics}: The ground truth consists of binary labels `Ambiguous' and `Unambiguous', which we regard as the positive and negative classes, respectively. It is imperative to recognize that any instance in which the model fails to discern ambiguities can have adverse implications during implementation. Consequently, we calculated precision, recall, and the F2 score across all experiments. We opted for the F2 score as the evaluation metric for this task, as we placed a heightened emphasis on minimizing false negatives, which could potentially culminate in penalties due to the model's failure to detect ambiguities, as opposed to false positives, where the model erroneously identifies ambiguities.

\textit{Retrieval-Augmented QA}: We engaged the non-legal stakeholders referenced earlier to validate the answers procured against a subset of generated CQs from the dataset. The answers were assessed for their accuracy in relation to the corresponding contract. An accurately retrieved answer was assigned a score of 1. When the LLM failed to retrieve an answer within the contract, a score of 0 was allocated. A score of 0 was also designated in scenarios where the generated answer was incorrect, ambiguous, or fabricated by the LLM. Each answer was evaluated by two annotators independently. 

\textit{Generation Quality of CQs}: In the absence of ground truth, we asked the annotators to manually evaluate the quality of the generated CQs. The annotators were not provided with the names of the LLMs or the techniques employed for generating the CQs. The annotators assessed each CQ based on a suite of pre-established individual and group-level criteria, as proposed by \citet{Zhang_2021}. The individual-level criteria were leveraged to evaluate each of the CQs generated for a sentence in isolation, while the group-level criteria were employed to assess the collective quality of the CQs generated for a given sentence. The individual-level parameters encompassed the following:

\begin{itemize}
\small
\itemsep0em 
    \item Grammaticality=0, if there is a syntax error or the generation result is not a question.
    \item Logicality=0, if the question is unclear or introduces additional ambiguities.
    \item Relevance=0 if the question asks for unambiguous information or does not directly address potential ambiguities in the contractual sentence provided.
    \item Specificity [score 0-4]: How specific is a question?
\end{itemize}

 For group-level evaluation, the parameters were:

\begin{itemize}
\small
\itemsep0em 
    \item Usefulness: No. of questions with any binary individual metric=0 and specificity score $<=2$.
    \item Redundancy: No. of questions that are semantically the same as any other generated question.
    \item Missing Cases [score 0-2]: The set of questions should cover most, if not all, of the significant attributes that require clarification. 
\end{itemize}

Two annotators independently scored each generated output for a given sentence. The scores were then averaged to calculate the aggregate score for all LLMs. 

Taking inspiration from prior work \cite{vicuna2023}, we also compared the generation quality of open-source LLMs by performing a relative comparison with ChatGPT using automated evaluation metrics. We adopted the evaluation protocol used by \citet{Zhang_2021} for this purpose. Our assessment incorporated Pairwise-BLEU, average-BLEU, and average-METEOR scores as metrics for the automated evaluation of CQGen. 

\vspace{-2mm}
\subsection{Implementation Details}
\label{s4.4}
We used Pytorch and Langchain to perform all our experiments. We incorporated the Pinecone library API to construct the vector database. For getting inferences using ChatGPT, we used the OpenAI API with the gpt-3.5-turbo engine. We used the Huggingface implementations of Alpaca-LORA, Vicuna, and Dolly-V2 \footnote{\href{https://huggingface.co/eachadea/vicuna-13b-1.1}{Vicuna}, \href{https://huggingface.co/tloen/alpaca-lora-7b}{Alpaca LORA}, \href{https://huggingface.co/databricks/dolly-v2-12b}{Dolly-V2}}. We set the temperature to 0 in all experiments to get deterministic results. The maximum new token length was set to 500. We divided the contract into chunks of 400 tokens to build the vector database with the intersection window as 20 tokens. We used OpenAI's text-embedding-ada-002 model for generating vectors. All our experiments were performed on an NVIDIA V100 GPU with 60 GB RAM and 32 GB GPU memory. We also note that all prompting experiments were conducted in the zero-shot setting due to the extensive context length and increased inference time needed for procuring few-shot results. 
\vspace{-2mm}
\section{Results and Analysis}
\label{s5}
\vspace{-1mm}
\begin{table}[t!]
\renewcommand{\arraystretch}{1.2}
\scriptsize
\centering
\begin{tabular}{ p{1.3cm} | p{2.5cm}  | p{0.6cm}  p{0.6cm}  p{0.6cm}  }
\hline
\textbf{Model} & \textbf{Prompting Technique} & \textbf{P} & \textbf{R} & \textbf{F2}\\
\hline
\multirow{5}{*}{ChatGPT} & Direct & 0.56 & 0.71 & 0.67 \\
                    
                         &  CoT	& 0.57 & 0.82 & 0.75 \\
                         
                         &  Modified CoT & 0.54 & 0.93 & 0.81 \\
                         
                         & ConRAP– Attribute Prompting & 0.54 & \textbf{0.98} & 0.84 \\
                         
                         & \textbf{ConRAP- Retrieval} & \textbf{0.64} & 0.97 & \textbf{0.87} \\
                        \hline
\multirow{5}{*}{Vicuna} & Direct & 0.55 & 0.11 & 0.13 \\
                         &  CoT	& 0.50 & 0.35 & 0.37 \\
                         &  Modified CoT & 0.54 & 0.64 & 0.61 \\
                         & ConRAP– Attribute Prompting & 0.53 & \textbf{0.97} & 0.83 \\
                         & \textbf{ConRAP- Retrieval} & \textbf{0.58} & 0.95 & \textbf{0.84} \\
                         \hline
\multirow{5}{*}{Alpaca LORA} & Direct & 0.64 & 0.12 & 0.14 \\
                         &  CoT	& 0.60 & 0.58 & 0.58 \\
                         &  Modified CoT & 0.59 & 0.60 & 0.60 \\
                         & ConRAP- Attribute Prompting & 0.55 & \textbf{0.83} & \textbf{0.75} \\
                         & \textbf{ConRAP- Retrieval} & \textbf{0.57} & 0.82 & \textbf{0.75} \\
                         \hline
\multirow{5}{*}{Dolly-V2} & Direct & 0.55 & 0.13 & 0.15 \\
                         &  CoT	& 0.55 & 0.26 & 0.29 \\
                         &  Modified CoT & 0.55 & 0.39 & 0.41 \\
                         & ConRAP– Attribute Prompting & 0.56 & 0.69 & 0.66 \\
                         & \textbf{ConRAP- Retrieval} & \textbf{0.61} & \textbf{0.76} & \textbf{0.72} \\
                         \hline
\end{tabular}
\vspace{-2mm}
\caption{Ambiguity Detection Results}
\label{table:1}
\end{table}

\begin{table}[t!]
\renewcommand{\arraystretch}{1.2}
\scriptsize
\centering
\begin{tabular}{ p{1.3cm} | p{2.5cm}  | p{0.6cm}  p{0.6cm}  p{0.6cm}  }
\hline
\textbf{Model} & \textbf{Prompting Technique} & \textbf{P} & \textbf{R} & \textbf{F2}\\
\hline
\multirow{3}{*}{ChatGPT} & Sentence-based Retrieval & 0.43 & 0.49 & 0.47 \\
                    
                         &  NER-based Retrieval	& 0.54 & 0.66 & 0.63 \\
                         
                         &  \textbf{ConRAP} & \textbf{0.64} & \textbf{0.97} & \textbf{0.87} \\
                        \hline
\multirow{3}{*}{Vicuna} & Sentence-based Retrieval & 0.48 & 0.39 & 0.40 \\
                         &  NER-based Retrieval	& 0.52 & 0.53 & 0.53 \\
                         &  \textbf{ConRAP} & \textbf{0.58} & \textbf{0.95} & \textbf{0.84} \\
                         \hline
\end{tabular}
\vspace{-2mm}
\caption{Ambiguity Detection using Retrieval-Augmented Prompts}
\label{table:2}
\vspace{-5mm}
\end{table}

\begin{table*}
 \renewcommand{\arraystretch}{1.2}
\centering
\scriptsize
\begin{tabular}{ p{1.5cm} | p{1cm} p{1cm}  p{1cm}  p{1cm}  p{1cm}  p{0.9cm}  p{1.1cm}  p{2cm}}
\hline
\textbf{Model}	& \textbf{\#CQs} &	\textbf{Grammar}	& \textbf{Relevant} &	\textbf{Logical}	& \textbf{Specific}	&	\textbf{Useful}	& \textbf{Redundant}	& \textbf{Missing} [0/1/2] \\
\hline
\textbf{ChatGPT}	& 250	& \textbf{1.0}	& \textbf{0.77}	& \textbf{0.98}	& \textbf{0.92}	& \textbf{0.70}	& \textbf{0.03}	& \textbf{0.83 / 0.16 / 0.01} \\
Vicuna	& 236	& 0.99	& 0.75	& 0.97  & 0.89 & 0.68 & 0.10 & 0.69 / 0.30 / 0.01 \\
Alpaca-LORA	& 453	& 0.93	& 0.58 & 0.92	& 0.73	& 0.27 &	0.13 &	0.6 / 0.34 / 0.06 \\
Dolly-V2	& 289	& 0.80	& 0.43	& 0.73	& 0.64 &	0.18 &	0.08 &	0.62 / 0.21/ 0.17 \\
\hline
\end{tabular}
\vspace{-2mm}
\caption{Human Evaluation Scores for  CQs generated using 100 randomly sampled sentences}
\label{table:3}
\vspace{-3mm}
\end{table*}

\begin{table}[!ht]
\renewcommand{\arraystretch}{1.2}
\scriptsize
\centering
\begin{tabular}{ p{1.5cm} | p{1cm}  p{1cm}  p{1.2cm}  }
\hline
\textbf{Model} & \textbf{Pairwise BLEU} & \textbf{Average BLEU} & \textbf{Average METEOR}\\
\hline
ChatGPT & 0.09 & - & - \\
\textbf{Vicuna} & \textbf{0.19} & \textbf{0.17} & \textbf{0.38} \\
Alpaca-LORA	& 0.26 & 0.14 & 0.35 \\
Dolly-V2 & 0.25	& 0.12 & 0.31 \\
\hline
\end{tabular}
\vspace{-2mm}
\caption{Automated Evaluation of open-source LLMs’ CQs against ChatGPT}
\label{table:4}
\vspace{-6mm}
\end{table}

\subsection{Answering RQs}
\label{s5.1}
\textbf{RQ1}: Table \ref{table:1} presents the results achieved using various zero-shot prompting strategies and LLMs on ambiguity detection. All results are reported on the dataset described in Section \ref{s4.1}. We report the results of both attribute prompting and retrieval modules separately. ConRAP's attribute prompting outperformed all other baselines across different LLMs. ChatGPT emerged as the most proficient, recording an F2 score of 0.84. Notably, there was a substantial enhancement in the F2 score by a margin of 20\% using Vicuna, whereas the margins of improvement stood at 16\% and 18\% using Alpaca-LORA and Dolly-V2, respectively. These findings underscore the efficacy of attribute prompting when deployed with open-source LLMs.

While attribute prompting considerably curtailed false negatives and improved the overall recall (0.98), it was hindered by a modest precision score (0.54). This outcome was anticipated, as the global context is not factored into attribute prompting. Consequently, attribute prompting yielded an abundance of CQs pertaining to references, keywords, and contract metadata. However, these CQs did not pinpoint the actual ambiguities in the contract.

We integrated the retrieval-augmented QA module into ConRAP, intending to filter out such CQs and enhance precision. Using ChatGPT, the retrieval-augmented QA module surpassed all benchmarks and elevated the precision from 0.54, as attained in the initial module, to 0.64. Correspondingly, the overall F2 score improved from 0.84 in the first module to 0.87. Similar gains were observed across all LLMs employed.

\textbf{RQ2}: We examine if addressing the local and global contexts independently within ConRAP improves ambiguity detection. The outcomes, using the top-performing LLMs ChatGPT and Vicuna, are presented. The marked disparity in the F2 score compared to the baselines, as documented in Table \ref{table:2} (approximately 25\%), substantiates that the separation between local and global contexts is crucial for improving ambiguity detection within contracts.

\textbf{RQ3}: Manual evaluation of question-answer pairs retrieved across a subset of 100 randomly sampled sentences revealed that ConRAP's Retrieval QA using ChatGPT performed with an accuracy of 83.8\%. The module correctly responded to 383 out of 457 CQs. Among other LLMs, the best-performing Vicuna answered 78\% CQs correctly. 

\textbf{RQ4}:  We compared the final set of CQs generated by open-source LLMs with those generated by ChatGPT. The findings are tabulated in Table \ref{table:4}. Among the LLMs employed, Vicuna emerged as the closest contender to ChatGPT in terms of the Pairwise-BLEU score, suggesting that the CQs generated by Vicuna exhibit greater diversity within each sentence as compared to Alpaca-LORA and Dolly-V2. It is pertinent to note that a lower Pairwise-BLEU score is indicative of increased diversity. Additionally, Vicuna surpassed the other LLMs in Avg-BLEU and METEOR scores. This was expected because Alpaca-LORA and Dolly-V2's responses are often ungrammatical and illogical.

The human evaluation results for generated CQs are presented in Table \ref{table:3}. Vicuna's generation quality closely approximates ChatGPT's across most of the parameters, with the exceptions being redundancy and missing cases, i.e., the scope of ambiguities that went undetected within each sentence. Vicuna demonstrated comprehensive ambiguity identification in approximately 69\% of the sentences and partial identification in roughly 30\%. In contrast, ChatGPT exhibited a more robust performance, comprehensively detecting ambiguities in 83\% of the sentences. This disparity can be attributed to ChatGPT's enhanced reasoning capabilities compared to open-source LLMs within a zero-shot setting. Notably, the CQs generated by ChatGPT and Vicuna were predominantly assessed as grammatically correct and logical. Furthermore, approximately 90\% and 75\% of the CQs were characterized as specific and relevant respectively. Alpaca-LORA and Dolly-V2 exhibit poor generation quality, with a mere 27\% and 18\% of the CQs, respectively, being deemed useful.
\vspace{-4mm}
\subsection{Analysis}
\label{s5.2}

\textit{Analyzing False Positives and False Negatives}: ConRAP experiences some instances of False Negatives (FNs), mainly when the generated CQs necessitate reasoning across the contract to ascertain the answer. Currently, the retrieval-QA within ConRAP does not incorporate document-level reasoning for answering CQs to minimize hallucinated responses. We plan to address this limitation in future work.

Despite an overall improvement in precision, False Positives (FPs) persist as a challenge for ConRAP. Common scenarios that contribute to FPs include: (1) Reliance on external knowledge for answering CQs, (2) Omission of information in publicly-released contracts for confidentiality reasons, (3) Trivial details not contributing to ambiguities within the sentence, and (4) Irrelevant CQs that can be addressed through commonsense knowledge and are not explicitly stated in contracts.  The distribution of FP and FN cases observed across the 500 questions evaluated with specific examples of each case are reported in Table \ref{table:5}. 

\begin{table*}
\renewcommand{\arraystretch}{1.2}
\scriptsize
\centering
\begin{tabular}{ p{5.2cm} p{3cm}  p{3cm}  p{2cm}  p{0.7cm}}
\hline
\textbf{Sentence} & \textbf{Generated CQ} & \textbf{Answer} & \textbf{Explanation} & \% \\
\hline
Neither Party shall have any claim under this Agreement or otherwise against the other Party for vacation pay, paid sick leave, retirement benefits, social security, or other employee benefits of any kind.	& What specific benefits are included in `other employee benefits of any kind'?	& Answer Not Specified	& Trivial ambiguities	& 29.9 \\
\hline
Notwithstanding anything to the contrary in this Agreement, if the Closing (as such term is defined in the Merger Agreement) does not occur, this Agreement shall be terminated, and the provisions herein shall have no force or effect.	& What is the definition of `Closing' in the Merger Agreement?	& Answer Not Specified	& Dependence on an external agreement	& 21.8 \\
\hline
Umbrella / Excess Liability coverage is inclusive of product liability with limits not less than \$5,000,000 per occurrence and aggregate.	& What is the maximum limit for the coverage in aggregate?	& \$5,000,000	& Incorrect Answer	& 20 \\
\hline
The upgrades will be available at the prices listed in the then-current price list as of the date of the Quote. &	Can you provide a copy of the then-current price list?	& Answer Not Specified	& Irrelevant CQs &	19.4 \\
\hline
[***] shall be [***] responsible for the costs of (a) any clinical trials conducted for purposes of obtaining Regulatory Approval for the Assay.	& Who or what does [***] refer to?	& Answer Not Specified	& Omitted information	& 8.9 \\
\hline
\end{tabular}
\vspace{-2mm}
\caption{Analyzing False Positives and False Negatives of ConRAP}
\label{table:5}
\vspace{-5mm}
\end{table*}

\textit{Inference Time}: The enhancement in the F2 score is accompanied by an increase in the time taken for the final inference. On a subset of 100 sentences randomly sampled from our dataset, the average inference time for ChatGPT using ConRAP was approximately 2.4 times longer than that with direct zero-shot prompting. This increase in inference time can be attributed to the need for a large set of tokens due to integrating two different modules within ConRAP.

\vspace{-2mm}
\section{Related Work}
\label{s6}
\vspace{-2mm}
\subsection{Clarification Question Generation}
\vspace{-1mm}
\label{s6.1}
CQGen is a well-studied problem in conversational systems literature with two primary use cases: (1) Conversational search and (2) Conversational QA \cite{rahmani2023survey}. \citet{kuhn2023clam} tackled the problem of ambiguous queries in conversational search by proposing using LMs for CQGen. \citet{wang2023zeroshot} proposed a zero-shot learning approach to enhance conversational search. \citet{rao-daume-iii-2018-learning} proposed the task of CQGen to identify missing information in the provided context. \citet{Zhang_2021} proposed a technique to improve the diversity and specificity of generated CQs by incorporating keywords. \citet{Kumar_2020} worked on ranking the generated CQs to provide valuable insights into their usefulness for information retrieval. No existing CQGen approach tackles retrieval from long and complex legal documents such as contracts. 
\vspace{-2mm}
\subsection{Ambiguity Detection in Contracts}
\label{s6.2}
 As contracts and other legal documents are rich sources of obligatory requirements, ambiguity detection in legal text is well-studied in Requirements Engineering (RE). \citet{massey2014} developed a taxonomy of ambiguity detection based on legal and linguistic understandings of ambiguities in regulatory requirements. Among the different types of ambiguities, efforts to detect Vagueness and Incompleteness in legal text have increased in the last decade \cite{bhatia-a, lebanoff, kotal}.

However, none of the previous approaches have incorporated global context to deal with ambiguity detection on the contract level. To the best of our knowledge, this is the first work that combines question generation and answering tasks to solve the issue of incorporating the entire contract document for ambiguity detection. Moreover, generating CQs to pinpoint contract ambiguities for assisting non-legal stakeholders has not been done before. 
\vspace{-2mm}

\section{Conclusion}
\label{s7}
\vspace{-1mm}
In this work, we introduced ConRAP, a retrieval-augmented prompting framework for generating CQs to pinpoint ambiguities in contractual sentences. ConRAP was assessed on document-level ambiguity detection and CQGen, utilizing sentences extracted and labeled from the CUAD dataset. The evaluation demonstrated that ConRAP, when used with ChatGPT, achieved an F2 score of 0.87 in ambiguity detection. Notably, 70\% of the CQs generated across a sample of 100 sentences were deemed useful by experts, and 83\% of ambiguities were comprehensively identified. In addition, ConRAP enhanced the performance across various open-source LLMs, with Vicuna exhibiting comparable results to ChatGPT on CQGen.

This work pioneers new paths for research in ambiguity detection and CQGen within contracts and marks the introduction of the CQGen task specific to contracts. Additionally, we make available a novel dataset designed for ambiguity detection, tailored to address ambiguities in the context of entire contract documents. Looking ahead, we aim to explore the degree of ambiguity levels in contractual sentences and investigate their influence on CQGen. Furthermore, we plan to integrate reasoning capabilities over contractual text to improve ambiguity detection.

\section{Limitations}

We acknowledge a few limitations of this work: (1) While efforts were made to ensure the representativeness of our dataset by incorporating 25 distinct categories of commercial contracts sourced from public filings with the U.S. Securities and Exchange Commission, the ability of ConRAP to identify ambiguities across a broader spectrum of legal documents accurately remains unverified. (2) While we experiment with different LLMs to analyze the results of ConRAP, the constantly evolving nature of ChatGPT may cause minor variations in the results obtained for ChatGPT-driven experiments over time. (3) The scalability of evaluating ambiguity detection and quality of the generated CQs is hindered by reliance on human annotations, requiring a significant amount of time and effort. It is also subject to evaluator biases. (4) ConRAP currently does not incorporate external or domain-specific knowledge that may be crucial for reasoning about ambiguities beyond the contractual text. Addressing this gap by integrating such knowledge sources is an area we have earmarked for future research. (5) ConRAP operates on the assumption that all vague, incomplete, or absent attributes within a contractual sentence bear equal significance. However, in real-world scenarios, the relative importance of these attributes may depend upon the intentions and priorities of the parties involved. In subsequent research, we intend to investigate how the interplay of different attributes influences the emergence of ambiguities in contracts. 6) Lastly, the efficiency of ConRAP is constrained by its inference time, which is adversely affected by dependencies on external APIs such as Pinecone. Improving the inference time is an area that warrants further optimization and refinement.

\section{Ethics Statement}

We recognize and acknowledge the potential for misuse associated with our proposed framework, including the possibility of adversarial exploitation to introduce ambiguities into contractual clauses. Additionally, there is a risk of unintended consequences due to inaccurate predictions when dealing with unseen data that can lead to a ripple effect on the dynamics of requirements elicitation and subsequent implementation of contractual terms. Consequently, we stress the importance of employing ConRAP judiciously and primarily as a supplementary tool before gathering project-specific requirements. It should not be considered a substitute for the critical expertise and scrutiny that non-legal stakeholders provide. Identifying ambiguities and generating CQs for contractual sentences are intended to serve as aids that can streamline and enhance the efficiency of experts, thereby reducing the time and effort expended in scrutinizing contractual clauses. It is crucial to ensure that ConRAP is aligned with ethical practices and is coupled with due diligence and expert consultation to prevent unintended ramifications.

\section{Data and Code Availability}

The dataset and code used in this work are publicly available \footnote{\href{https://github.com/anmolsinghal98/Clarification-Question-Generation-for-Contracts}{GitHub Repository}}. 

\nocite{*}
\section{Bibliographical References}\label{sec:reference}

\bibliographystyle{lrec-coling2024-natbib}
\bibliography{paper}

\section{Language Resource References}
\label{lr:ref}
\bibliographystylelanguageresource{lrec-coling2024-natbib}
\bibliographylanguageresource{languageresource}

\appendix

\section{Examples of Ambiguity Types}
\label{AppendixA}
\textit{Vagueness}: The contractual sentence “Data Processor shall keep Personal Information sufficiently isolated from other data on the server in an appropriate manner to prevent it from being misused.” is vague because it includes vague phrases such as `sufficiently isolated' and `appropriate manner’ which do not have any clear boundaries of meaning.

\textit{Incompleteness}:  The contractual sentence “The Service Provider shall notify the Customer in writing of any changes to the Service Level Agreement.” is incomplete because it does not specify the deadline by which the service provider should notify the customer, leaving it open to interpretation.

\textit{Referential}: The contractual sentence “The Service Provider shall provide support to the Customer according to the applicable provisions in the Agreement” contains an ambiguous reference to the provisions of the Agreement. 

\section{Attribute Prompt List}
\label{AppendixB}
The attributes that typically occur in contractual sentences, as identified in this work, can be categorized as follows:

\begin{itemize}
\vspace{-2mm}
\itemsep0em 
\small
    \item Actor - Who is supposed to act?
    \item Action to be performed - What is to be done?
    \item Object Information / Definitions - What is being acted upon?
    \item Condition - Under what condition or circumstance is the action performed?
    \item Modality / Intent of Action - How certain is the occurrence of the action?
    \item Quantitative Attributes - How much, to what extent?
    \item Time / Duration / Frequency of action / Leadtime– When / How long / How often?
    \item Location / Jurisdiction of action – Where?
    \item Qualitative Attributes – How?
    \item References – With respect to? / According to?
\end{itemize}

Since most contract documents are written repetitively, we expect the identified types of attributes to apply to contracts outside the scope of our dataset. An attribute of an unspecified type, if encountered, can be mapped to an already identified type closest to it in terms of semantic understanding. For instance, an attribute of an unspecified type influencing the object in the sentence can be mapped to `object information'.

We also note that we did not include the attribute list in the final prompt of the ConRAP framework. While designing the final prompt, we also experimented with the setting where the attribute list is included within the prompt. However, we found that including the attribute list did not improve the results and was computationally more expensive than using our final prompt due to the increased length of the input prompt. 

\section{Question-Answering Prompt}
\label{AppendixC}
\vspace{2mm}

\noindent\fbox{%
\small
\renewcommand{\arraystretch}{1.4}
    \parbox{7.5cm}{%
Instructions: Use the following pieces of context to answer a question related to a contractual sentence at the end. If you don’t know the answer, respond ‘Answer Not Specified’. Do not try to make up an answer. \\
Context: \textit{$<$retrieved chunks$>$} \\
Question: LLM generated CQ \\
Answer: 
    }%
}
\vspace{2mm}

\section{Prompting Baselines}
\label{AppendixD}
The prompts corresponding to Direct, CoT, and modified CoT are as follows:

\subsection{Direct Prompting}

Instructions: Predict if the given contractual sentence is ambiguous and generate a list of clarification questions that need to be asked to stakeholders to resolve ambiguities.
\\
Input: \textit{$<$user input$>$} 
\\
Output: \\
Prediction: \textit{$<$Ambiguous / Not Ambiguous$>$} \\
Clarification Questions: []

\subsection{CoT Prompting}

Instructions: Reason step by step if the given contractual sentence is ambiguous and generate a list of clarification questions that need to be asked to stakeholders to resolve ambiguities.
\\
Input: \textit{$<$user input$>$} 
\\
Output: \\
Prediction: \textit{$<$Ambiguous / Not Ambiguous$>$} \\
Reason for Ambiguity: \\
Clarification Questions: []

\subsection{Modified CoT Prompting}

Instructions: Reason step by step if the given contractual sentence is ambiguous and generate a list of clarification questions that need to be asked to stakeholders to resolve ambiguities. A sentence is ambiguous when it contains semantic elements that are vague, incomplete, or missing. Even minor ambiguities should be identified. 
\\
Input: \textit{$<$user input$>$} 
\\
Output: \\
Prediction: \textit{$<$Ambiguous / Not Ambiguous$>$} \\
Reason for Ambiguity:  \\
Clarification Questions: []

\end{document}